\documentclass[conference]{IEEEtran}
\IEEEoverridecommandlockouts
\usepackage{cite}
\usepackage{amsmath,amssymb,amsfonts}
\usepackage{url} 
\usepackage{algorithm}
\usepackage{algpseudocode}
\usepackage{booktabs} 
\usepackage{booktabs}    
\usepackage{makecell}  
\usepackage{tikz}
\usetikzlibrary{positioning, fit, arrows.meta, backgrounds, calc}
\usepackage{graphicx}
\usepackage{textcomp}
\usepackage{xcolor}
\usepackage{subfig}
\usepackage{todonotes}
\usepackage{comment}
\def\BibTeX{{\rm B\kern-.05em{\sc i\kern-.025em b}\kern-.08em
    T\kern-.1667em\lower.7ex\hbox{E}\kern-.125emX}}
\begin{document}

\title{SurvCF(t): Counterfactual Explanations for Survival Analysis in Predictive Maintenance Multivariate Time Series Data\\
}

\author{\IEEEauthorblockN{1\textsuperscript{st} Zara Karazian}
\IEEEauthorblockA{\textit{Department of Computer and Systems Sciences (DSV)} \\
\textit{Stockholm University}\\
Stockholm, Sweden \\
zara.karazian@dsv.su.se}
\and
\IEEEauthorblockN{2\textsuperscript{nd} Panagiotis Papapetrou}
\IEEEauthorblockA{\textit{Department of Computer and Systems Sciences (DSV)} \\
\textit{Stockholm University)}\\
Stockholm, Sweden \\
panagiotis@dsv.su.se}
\and
\IEEEauthorblockN{3\textsuperscript{rd} Sindri Magnússon}
\IEEEauthorblockA{\textit{Department of Computer and Systems Sciences (DSV)} \\
\textit{Stockholm University)}\\
Stockholm, Sweden \\
sindri.magnusson@dsv.su.se}
\and
\IEEEauthorblockN{4\textsuperscript{th} Erik Frisk}
\IEEEauthorblockA{\textit{Department of Electrical Engineering (ISY)} \\
\textit{Linköping University)}\\
Linköping, Sweden \\
erik.frisk@liu.se}
\and
\IEEEauthorblockN{5\textsuperscript{th} Tony Lindgren}
\IEEEauthorblockA{\textit{Department of Computer and Systems Sciences (DSV)} \\
\textit{Stockholm University)}\\
Stockholm, Sweden \\
tony@dsv.su.se}
}

\maketitle

\begin{abstract}

Predictive maintenance relies on accurate Remaining Useful Life estimation, often formulated using survival analysis over multivariate time-series data. While modern deep survival models achieve strong predictive performance, their black-box nature limits their use in safety-critical settings where actionable insight is required. In this work, we introduce \textit{SurvCF(t)}, the first framework for generating counterfactual explanations for survival models operating on time-series data. \textit{SurvCF(t)} identifies minimal, plausible, and temporally consistent changes to an asset’s operational history that increase its predicted life time, framing explanation as a constrained optimization problem combining validity, proximity, sparsity, and plausibility. We evaluate the method on multiple benchmarks, including C-MAPSS, N-CMAPSS, and a real-world case study of the Scania Component\_X dataset, demonstrating its ability to produce actionable and interpretable interventions. Our results show that \textit{SurvCF(t)} bridges the gap between survival prediction and prescriptive maintenance, enabling explainable and decision-oriented AI for maintenance strategies. 
\end{abstract}


\section{Introduction}

Modern industrial fleets produce continuous, high‑frequency telemetry from on‑board sensors: temperatures, pressures, flows, vibrations, electrical signals, and control commands. In vehicular applications (buses, trucks, trains), these streams are used to monitor each vehicle during its duty cycle so operators can deliver reliable service while minimizing cost and downtime \cite{lee2014service}. Maintenance practice in industry has historically fallen into three broad paradigms \cite{nunes2023challenges,kharazian2025challenges}. Reactive maintenance repairs equipment only after failure and is simple but often unacceptable for safety‑critical fleets. Preventive maintenance schedules servicing at fixed intervals (calendar‑ or usage‑based), reducing unexpected failures but wasting resources when parts are replaced early. Predictive (condition‑based) maintenance uses live telemetry and data analytics to estimate asset health and intervene only when needed, aiming to minimize both unplanned downtime and unnecessary work \cite{jardine2006review,mobley2002introduction,elwany2008sensor}.

At the heart of predictive maintenance (PdM) lies Remaining Useful Life (RUL) estimation: given the observed operational history of an asset up to the current time, how much longer can it be expected to operate before reaching a failure threshold or requiring maintenance \cite{si2011remaining,lei2018machinery,banjevic2009remaining}? Accurate RUL forecasts allow planners to schedule interventions with sufficient lead time, prioritize scarce repair capacity, and optimize spare‑parts  \cite{rahat2026contrastive}.


Real PdM data present some challenges that complicate RUL modeling \cite{kharazian2025challenges}. First, the inputs are multivariate time series: many sensor channels that evolve jointly over time and whose informative signal resides in temporal patterns (trends, drifts, transient events) rather than single snapshots. Second, measurements are noisy, correlated, and often sampled irregularly in the field, which complicates temporal aggregation. Third, industrial assets are engineered for high reliability: failures are relatively rare, and many observations end without an observed failure (the asset is still operating when the recording stops). Consequently, most datasets are heavily \emph{censored} and contain few run‑to‑failure examples. Naïvely treating censored records as failures (or discarding them) both wastes data and biases estimators. These issues make standard supervised learning insufficient and motivate survival‑statistical approaches that natively handle censoring and model the time‑to‑event distribution \cite{jenkins2005survival,klein2003survival}.
Survival analysis naturally incorporate censored observations in the study. Recent work has extended deep learning to survival problems, enabling models (e.g., recurrent or transformer architectures) to learn complex temporal dependencies from high‑dimensional sensor windows and often improving RUL accuracy over classical approaches \cite{katzman2018deepsurv,lee2018deephit}. However, higher predictive performance alone does not satisfy operational needs. Maintenance decisions are prescriptive: planners must decide what to change (engine setpoints, duty cycle, load, temporary operational constraints) so that a vehicle can safely complete a mission. Black‑box predictors explain that a failure is likely but not \emph{how} to avoid it.

Explainable AI methods (attribution, attention) help diagnose model predictions by highlighting influential sensors and time periods, but they remain descriptive \cite{molnar2019interpretable}. The decisive question for PdM is prescriptive: given the most recent observed window, which minimal and feasible adjustments to the actionable controls would increase the predicted lifetime sufficiently so the vehicle can finish its duty? Counterfactual explanations answer precisely this type of question by proposing concrete changes to inputs that lead to a preferred outcome \cite{wachter2017counterfactual}.


\subsection{Motivation}
The motivation for this work comes from PdM in the vehicular industry, where one of the main operational goals is to ensure that a vehicle can complete its assigned duty without failing on the road. In this setting, it is not sufficient to estimate RUL alone, but also, once a risk of failure is detected, the system should provide actionable guidance on how to adjust current operating conditions to extend RUL. 
This is especially important when the decision is made from the latest observed window of sensor measurements, because the recommendations must be grounded in the vehicle's current state and anticipate what should be changed in the future to improve health conditions. A counterfactual explanation system is therefore attractive. It can estimate the RUL based on the current usage pattern and suggest concrete adjustments to the actionable variables so that the vehicle can complete its duty safely and return to the workshop for maintenance, rather than failing in operation.

\begin{figure}[htbp]
\centering
\includegraphics[width=0.8\columnwidth]{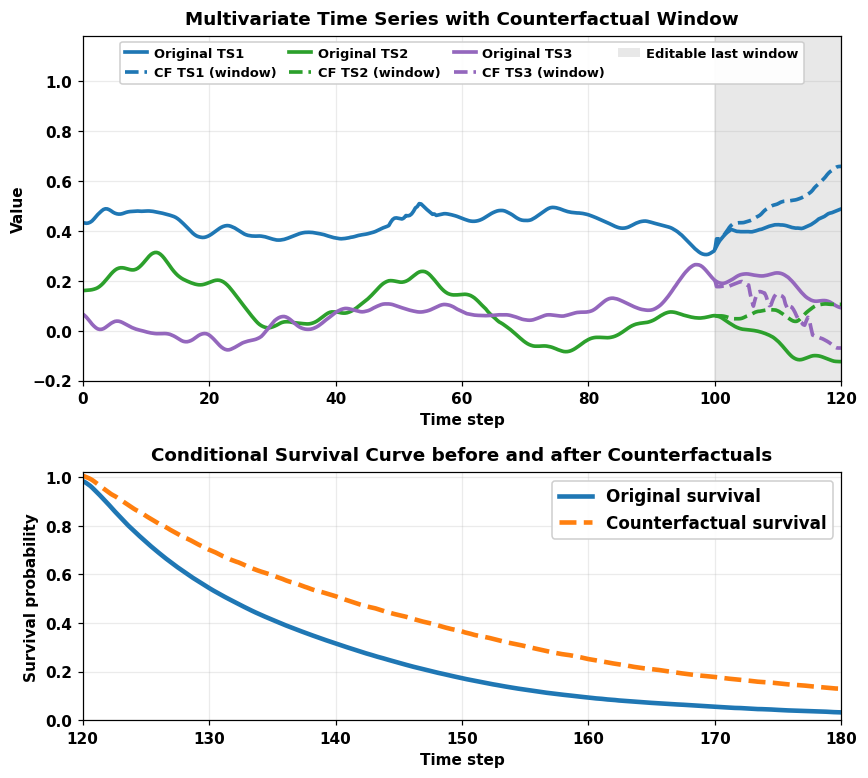}
\caption{Overview of SurvCF(t): Given the latest observed window of multivariate sensor measurements, SurvCF(t) generates a counterfactual trajectory by modifying only actionable variables. The resulting counterfactual leads to an improved predicted survival curve and increased RUL, providing actionable guidance for predictive maintenance decisions.}
\label{idea}
\end{figure}

Fig.~\ref{idea} illustrates the idea behind this work. Given the latest observed window of multivariate sensor readings, \textit{SurvCF} uses the vehicle's current operating condition to generate a counterfactual adjustment to the editable features, shifting the predicted survival curve upward and corresponding to a larger RUL. 

\subsection{Our Contributions}

Despite recent progress in both survival modeling and counterfactual explanations on multivariate time series data, their intersection remains largely unexplored. In particular, to the best of our knowledge, \textbf{no prior work has developed counterfactual explanations for survival models operating on multivariate time-series data}. Existing approaches either focus on static tabular survival settings or generate counterfactuals for classification and regression models, without accounting for the temporal structure and censoring inherent to PdM data. This gap is critical, as real-world maintenance decisions rely on sequential observations and require actionable, temporally consistent interventions.

To address this limitation, we propose \textit{SurvCF(t)}, a novel framework for generating counterfactual explanations for survival models operating on multivariate time series data in PdM. \textit{SurvCF(t)} bridges the gap between accurate time-to-event prediction and actionable decision support by producing minimal, plausible, and physically meaningful interventions that improve predicted RUL.
We can summarize our contributions as follows:

\begin{itemize}
\item \textbf{Counterfactual explanations for survival-based predictive maintenance.}
We present a novel framework for generating counterfactual explanations for survival models operating on multivariate time-series data, enabling the analysis of how changes in an asset's operational history affect its predicted remaining useful life.
\item \textbf{Survival-aware counterfactual optimization.}
We formulate counterfactual generation as a constrained optimization problem that seeks improved predicted RUL while balancing proximity to the original instance, sparsity of changes, and plausibility of the resulting trajectory.
\item \textbf{Evaluation on multiple predictive-maintenance benchmarks.}
We assess the proposed framework on several datasets with different asset types and operating conditions, including C-MAPSS, N-CMAPSS, and Scania Component\_X, and analyze the quality of the generated explanations in terms of validity, sparsity, proximity, and plausibility.
\end{itemize}

\section{Background and Related Works}

\subsection{Counterfactual Explanations for Time Series}

Counterfactual explanations have emerged as a principled approach in explainable artificial intelligence for providing actionable insight into model predictions. Rather than attributing importance to features, they answer a contrastive question: what minimal change to the input would lead to a different, desired outcome. Early work in this area, notably by Wachter et al. \cite{wachter2017counterfactual}, formulated counterfactual generation as an optimization problem that balances altering the prediction with maintaining proximity to the original instance. This formulation has since become the foundation for a large body of work in tabular settings, where inputs are assumed to be independent and unordered, as summarized in~\cite{guidotti2024counterfactual,verma2024counterfactual}.

Extending counterfactual reasoning to time series is significantly more challenging. Sequential data exhibit temporal dependencies, smoothness constraints, and cross-variable interactions that must be preserved to produce realistic explanations. Treating a time series as a flat vector often leads to implausible or discontinuous counterfactuals, motivating methods that explicitly incorporate temporal structure. Existing approaches can be broadly categorized into three groups: instance-based methods, which construct counterfactuals from observed examples to ensure plausibility (e.g., NUN \cite{dasarathy1995nearest}, CoMTE \cite{ates2021counterfactual}, MASCOTS \cite{pludowski2025mascots}); optimization-based methods, which extend classical formulations with temporal regularization (e.g., TimeX \cite{filali2022mining}, AB-CF \cite{li2023attention}, Glacier \cite{wang2024glacier}, LORE adaptations \cite{lampridis2020explaining}); and representation- or subsequence-based approaches, which modify salient temporal regions or operate in latent spaces to preserve coherence (e.g., DiscoX \cite{bahri2024discord}, CELS \cite{li2023cels}, M-CELS \cite{li2024m}).

Across these approaches, the central challenge is to generate counterfactuals that are not only valid but also sparse, proximal, and plausible under the constraints of the data-generating process. This is particularly important in domains such as PdM, where variables are physically coupled and recommended changes must correspond to feasible interventions. The challenge is further amplified in survival analysis settings, where the outcome is a time-dependent quantity rather than a single prediction, requiring counterfactuals to reason about changes in the evolution of risk over time.

\subsection{Survival Analysis and Remaining Useful Life}

Survival analysis is a statistical framework for modeling the time until an event of interest occurs, with roots in biostatistics and reliability engineering~\cite{kalbfleisch2002statistical}. Typical events include failure, degradation beyond an acceptable threshold, or transition between operational states. A defining feature of this setting is that event times are often only partially observed. Such observations are referred to as \emph{censored} and are typically classified as right censoring (the event has not occurred by the end of observation), left censoring (the event occurred before observation began), or interval censoring (the event is known only to lie within a time interval).

In many applications, and especially in predictive maintenance, right censoring is the dominant case. Modern industrial systems are designed for high reliability, and observation windows are often limited by practical constraints such as data collection periods or maintenance schedules. As a result, many units do not fail during the observation period, and their true failure times remain unknown. Survival analysis explicitly accounts for these partially observed lifetimes, allowing all available data to contribute to model estimation, in contrast to standard regression approaches that require fully observed outcomes.

Formally, in the right-censored setting the data are given by $\{(\mathbf{x}^{(i)}, T^{(i)}, \delta^{(i)})\}_{i=1}^{N}$, where $\mathbf{x}^{(i)}$ represents observed covariates (e.g., sensor measurements or time series), $T^{(i)} \in \mathbb{R}_+$ is the observed time, and $\delta^{(i)} \in \{0,1\}$ indicates whether the event is observed ($\delta^{(i)}=1$) or censored ($\delta^{(i)}=0$). Survival models describe the distribution of the event time through time-dependent quantities such as:

\begin{itemize}
    \item \textbf{Survival function} $S(t) = \mathbb{P}(T > t)$: the probability of surviving beyond time $t$.

    \item \textbf{Probability density function (PDF)} $f(t) = h(t)\,S(t)$: the probability density of an event occurring at time $t$.
    
    \item \textbf{Hazard function} $h(t)=\dfrac{f(t)}{S(t)}$: the instantaneous risk of experiencing the event at time $t$, given survival up to $t$.
    
    \item \textbf{Expected remaining lifetime} $m(t) = \mathbb{E}[T - t \mid T > t]$, commonly referred to as \emph{Remaining Useful Life} (RUL) in engineering applications.
\end{itemize}

When applied to predictive maintenance, survival analysis provides a natural probabilistic framework for estimating RUL from observed system behavior \cite{voronov2018data,rahat2023bridging,kargar2024shap,rahat2024survloss}. Early approaches typically rely on tabular representations with engineered features and employ models such as the Cox proportional hazards model or parametric lifetime distributions (e.g., Weibull), from which RUL is derived via the estimated survival function. More recent work extends these models to time-series data by incorporating time-dependent covariates or combining survival objectives with neural networks. Deep survival models, including neural Cox variants (e.g., DeepSurv\cite{katzman2018deepsurv}.) and architectures such as DeepHit \cite{lee2018deephit}, learn representations directly from sequential sensor data and capture both temporal dependencies and uncertainty in failure times.

\subsection{Counterfactual Explanations in Survival Analysis}

Compared to classification and regression settings, counterfactual (CF) explanations for survival models remain relatively underexplored. A small but growing body of work has begun to investigate how counterfactual reasoning can be adapted to time-to-event data, where predictions are expressed as survival functions or risk over time rather than discrete outputs.

Several works approach this problem from a causal or representation learning perspective. Chapfuwa et al. \cite{chapfuwa2021enabling} and Gupta et al. \cite{gupta2023deep} propose learning balanced representations to enable counterfactual reasoning under treatment effects and censored outcomes. Similarly, Nagpal et al. \cite{nagpal2022counterfactual} introduce counterfactual phenotyping for censored time-to-event data, focusing on identifying subgroups with different risk profiles. While these approaches provide important theoretical foundations, their primary objective is causal inference rather than generating actionable, instance-level explanations for model predictions.

Other works focus more directly on explaining survival models. Kovalev et al. \cite{kovalev2021counterfactual} and Alabdallah et al. \cite{alabdallah2024understanding} propose methods for generating counterfactual explanations that alter predicted survival outcomes, while Tran et al. \cite{tran2022decision} explore their use in decision support applications. However, these methods are restricted to tabular settings and operate on static covariates, without explicitly modeling temporal dynamics or sequential data. As a result, they do not address the key challenges associated with time-dependent inputs, such as temporal consistency, smoothness, and cross-time dependencies.

Existing approaches demonstrate the potential of counterfactual explanations in survival analysis but remain limited in scope. In particular, to the best of our knowledge, no prior work has developed counterfactual explanations for survival models operating on multivariate time-series data. This gap is critical in domains such as predictive maintenance, healthcare monitoring, and other longitudinal settings, where system behavior evolves over time and decisions must be based on sequential observations. Addressing this limitation constitutes a central contribution of this work.

\section{Problem Formulation and Algorithm}

We consider the problem of generating counterfactual explanations for survival models operating on multivariate time-series data. The goal is not only to predict the time-to-event, but also to provide actionable insights into how the predicted outcome can be improved. To this end, we propose a two-stage framework consisting of: (i) a survival model that estimates time-dependent risk from observed system trajectories, and (ii) a counterfactual generator that produces minimally perturbed input sequences that increases the predicted time-to-event while remaining sparse, close to the original, and plausible under the training distribution.

Let $\mathcal{D} = \{(X^{(i)}, T^{(i)}, \delta^{(i)})\}_{i=1}^{N}$
denote a dataset of $N$ independent units. For each unit $i$:

\begin{itemize}
    \item $X^{(i)} \in \mathbb{R}^{m_i \times D}$ is a multivariate
    signal with $m_i$ timesteps along the rows and $D$ feature
    channels along the columns.
    \item $T^{(i)} \in \mathbb{R}_+$ is the observed
    survival time, $T^{(i)} = \min\bigl(T^{(i)}_\star, C^{(i)}\bigr)$,
    where $T^{(i)}_\star$ is the true failure
    time and $C^{(i)}$ is the censoring time.
    \item $\delta^{(i)} \in \{0,1\}$ is the event indicator:
    $\delta^{(i)} = 1$ if failure was observed
    ($T^{(i)}_\star \leq C^{(i)}$),
    $\delta^{(i)} = 0$ if the unit was right-censored.
\end{itemize}

We partition the feature index set $\{1,\ldots,D\}$ into two
disjoint subsets with
$\mathcal{F}_e \cup \mathcal{F}_f = \{1,\ldots,D\}$ and
$\mathcal{F}_e \cap \mathcal{F}_f = \emptyset$, where
$\mathcal{F}_e$ collects the editable (actionable) features the
counterfactual may modify, and $\mathcal{F}_f$ the fixed features
(e.g.\ environmental or non-controllable variables) that the
counterfactual must leave unchanged.



The goal is to learn (i) a survival model $f_\theta$ that estimates the conditional hazard function given the recent history of $X$, and (ii) a counterfactual operator $g$ that maps an input window to a minimally perturbed counterfactual window for which $f_\theta$ predicts a longer remaining time-to-event.

\subsection{Survival Modeling}\label{sec:survival modeling}

We model survival from multivariate time-series input using a recurrent neural network with a discrete-time hazard head. The network reads a fixed-length window of recent observations and outputs a conditional hazard probability for each of $T_{\max}$ future time bins. The training pipeline has three components: a windowing procedure that converts variable-length unit trajectories into fixed-length samples (Sec.~\ref{sec:windowing}), a discrete-time hazard model that maps each window to a hazard vector (Sec.~\ref{sec:hazard-model}), and a censored log-likelihood objective that accommodates both observed events and right-censored windows (Sec.~\ref{sec:training-loss}).
Algorithm \ref{alg:train} summarizes all the steps on survival modeling of SurvCF.

\vspace{2mm}
\subsubsection{Window-Based Representation}
\label{sec:windowing}
For each unit $i$, we extract windows of length $L$ by sliding backward
from the end of the trajectory in strides of $L$, yielding samples
\begin{equation}
\bigl(X^{(i,k)},\, t^{(i,k)},\, \delta^{(i,k)}\bigr), \quad
X^{(i,k)} \in \mathbb{R}^{L \times D},
\end{equation}
where $k$ indexes the window's distance from the end-of-trajectory
($k = 0$ is the most recent window). The discrete-time index
$t^{(i,k)} = \lfloor k L / \Delta t \rfloor$ encodes how many future
hazard bins of width $\Delta t$ remain until the trajectory end. The
event indicator $\delta^{(i,k)}$ equals $\delta^{(i)}$ only at $k = 0$
and is set to $0$ otherwise, encoding the assumption that earlier
windows are observed-but-not-yet-failed.


\vspace{2mm}

\subsubsection{Discrete-Time Hazard Model}
\label{sec:hazard-model}
Following standard discrete survival analysis, we partition the future
time axis into $T_{\max}$ bins of width $\Delta t$ and model the
conditional hazard
$h_t = \Pr(\text{event in bin } t \mid \text{survived to } t{-}1)$
for each bin. A long short-term memory (LSTM) network $\phi_\theta$
followed by an elementwise sigmoid maps a window
$X \in \mathbb{R}^{L \times D}$ to a hazard vector
\begin{equation}
h(X) = \sigma\bigl(\phi_\theta(X)\bigr) \in (0, 1)^{T_{\max}}.
\end{equation}
The conditional survival function and expected remaining time-to-event
follow as
\begin{equation}
S_t(X) = \prod_{s<t}\bigl(1 - h_s(X)\bigr), \qquad
\widehat{\text{RUL}}(X) = \Delta t \sum_{t=0}^{T_{\max}-1} S_t(X).
\end{equation}

\subsubsection{Training Objective}
\label{sec:training-loss}
For a single windowed sample $(X^{(n)}, t^{(n)}, \delta^{(n)})$, the
discrete-time censored log-likelihood contribution is
\begin{equation}
\begin{split}
\ell^{(n)}(\theta) ={}& \delta^{(n)}\!\bigl[\log S_{t^{(n)}}(X^{(n)}) + \log h_{t^{(n)}}(X^{(n)})\bigr] \\
& + (1-\delta^{(n)})\,\log S_{t^{(n)}+1}(X^{(n)}).
\end{split}
\end{equation}
The first term applies to observed events: it rewards a high hazard at
the event bin together with high survival up to that bin. The second
term applies to right-censored samples and rewards survival past the
censoring bin. Parameters $\theta$ are estimated by minimizing the
aggregate negative log-likelihood,
\begin{equation}
\mathcal{L}_{\text{surv}}(\theta) = -\sum_{n=1}^{N_w} \ell^{(n)}(\theta),
\end{equation}
where $N_w$ is the total number of training windows. Optimization uses
Adam with gradient clipping at norm $1$ to mitigate exploding gradients
common in recurrent encoders. To prevent leakage between windows of the same unit, model selection
uses GroupKFold cross-validation with $K=5$ folds and engines treated
as groups. All windows from a given engine are kept together in
either the training or validation split of each fold. The final
model is the fold-wise checkpoint with the lowest validation loss.

\begin{algorithm}[t]
\caption{Survival Model Training}
\label{alg:train}
\begin{algorithmic}[1]
\Require Dataset $\mathcal{D}$; window length $L$; folds $K$; epochs $E$
\Ensure Trained model $f_{\theta^\star}$

\Statex \textbf{(1) Window-Based Representation}
\State $\mathcal{W} \gets$ \Call{SlidingWindows}{$\mathcal{D}, L$}

\Statex \textbf{(2) Discrete-Time Hazard Model}
\Statex \quad $f_\theta : X \mapsto h(X) \in (0,1)^{T_{\max}}$,\; loss $\mathcal{L}_{\text{surv}}(\theta;\mathcal{W})$

\Statex \textbf{(3) Training Objective with GroupKFold CV}
\State $\mathcal{L}_{\text{best}} \gets +\infty$
\For{fold $j \in$ \Call{GroupKFold}{$\mathcal{W}, K$}}
  \State Initialize $\theta$
  \For{$e = 1, \ldots, E$}
    \State $\theta \gets$ Adam step on $\mathcal{L}_{\text{surv}}$ over fold-$j$ train
    \State $\mathcal{L}_{\text{val}}^{(e)} \gets \mathcal{L}_{\text{surv}}$ on fold-$j$ val
  \EndFor
  \State Restore $\theta$ to $e^\star = \arg\min_e \mathcal{L}_{\text{val}}^{(e)}$ \Comment{best-within-fold}
  \If{$\mathcal{L}_{\text{val}}^{(e^\star)} < \mathcal{L}_{\text{best}}$}
    \State $\mathcal{L}_{\text{best}} \gets \mathcal{L}_{\text{val}}^{(e^\star)}$;\; $\theta^\star \gets \theta$
  \EndIf
\EndFor
\State \Return $f_{\theta^\star}$
\end{algorithmic}
\end{algorithm}

\subsection{Counterfactual Generation}
\label{sec:cf}
Having trained the survival model in Section~\ref{sec:survival modeling}, we now
address the inverse question central to this work: for a given test
engine, what minimal and realistic modification of its editable input
features would extend its predicted remaining useful life? The
counterfactual generation procedure below answers this by perturbing
the input window so that the model's RUL prediction crosses a
user-specified improvement target, while keeping the perturbation
close to the original, sparse in the number of altered features, and
consistent with the training distribution. 
Fig.~\ref{idea} illustrates the SurvCF idea: by perturbing only the last window of a multivariate input (top), the model's predicted survival curve shifts upward (bottom), indicating a longer RUL.

For a test window $X \in \mathbb{R}^{L \times D}$ with predicted survival
$\widehat{\text{RUL}}(X)$, a counterfactual $X' \in \mathbb{R}^{L \times D}$
is a perturbed input differing from $X$ only on editable features
($X'_{:,f} = X_{:,f}$ for all $f \in \mathcal{F}_f$).

We parameterize the perturbation as $Z \in \mathbb{R}^{L \times D}$,
initialized at $0$, and define a projected counterfactual
\begin{equation}
\label{eq:cf-proj}
X'(Z) = \mathrm{clip}\bigl(X + Z \odot M_e,\; \mathbf{lb},\; \mathbf{ub}\bigr),
\end{equation}
where $M_e \in \{0,1\}^{D}$ is a binary mask zeroing out non-editable
features and $(\mathbf{lb}, \mathbf{ub})$ are per-feature box constraints
set to the $\{q,\,1{-}q\}$ quantiles of the training distribution. We
write $\boldsymbol{\eta} := X'(Z) - X$ for the resulting perturbation.

We adopt the following objective function for CF generation, minimized by
Adam:
\begin{equation}
\label{eq:cf-loss}
\begin{split}
\mathcal{L}_{\text{CF}}(Z) ={}& \lambda_{\text{rul}}\bigl[\text{RUL}^\star_{\text{int}} - \widehat{\text{RUL}}(X+\boldsymbol{\eta})\bigr]_+ \\
& + \lambda_{\text{dist}}\,\lVert\boldsymbol{\eta}\rVert_2 + \lambda_{\text{smooth}}\,\lVert D_t\,\boldsymbol{\eta}\rVert_2 \\
& + \lambda_{\text{plaus}}\,\bigl[d_M(X+\boldsymbol{\eta}) - d_M(X)\bigr]_+,
\end{split}
\end{equation}
where $[\cdot]_+$ denotes the ReLU and
$\lambda_{\text{rul}}, \lambda_{\text{dist}}, \lambda_{\text{smooth}},
\lambda_{\text{plaus}} \geq 0$ are scalar weights. Each term in
Eq.~\eqref{eq:cf-loss} supports one of the desirable CF properties from
the literature \cite{wachter2017counterfactual, guidotti2024counterfactual,verma2024counterfactual}:

\begin{itemize}
    \item \textbf{Validity} is supported by the first term:
    $$\lambda_{\text{rul}}\,\bigl[\text{RUL}^\star_{\text{int}} - \widehat{\text{RUL}}(X+\boldsymbol{\eta})\bigr]_+ .$$
    Given a user-chosen relative improvement $\rho \in (0,1]$, the
    \emph{nominal target} $\text{RUL}^\star = (1+\rho)\widehat{\text{RUL}}(X)$
    is the RUL we want the counterfactual to achieve. To create headroom
    for the subsequent sparsification step, the hinge loss aims at an
    \emph{internal target} $\text{RUL}^\star_{\text{int}} = \kappa\,\text{RUL}^\star$,
    where $\kappa > 1$ is a fixed overshoot factor. The ReLU activates
    only while $\widehat{\text{RUL}}$ falls short, so the gradient
    pushes $\boldsymbol{\eta}$ in directions that increase the predicted
    RUL.

    \item \textbf{Proximity} is supported by the second and third terms:
    $$\lambda_{\text{dist}}\,\lVert\boldsymbol{\eta}\rVert_2 \;+\; \lambda_{\text{smooth}}\,\lVert D_t\,\boldsymbol{\eta}\rVert_2 .$$
    The $\ell_2$ term keeps the perturbation magnitude small, while the
    smoothness term discourages high-frequency cycle-to-cycle jumps. Here
    $D_t$ denotes the discrete first-difference operator along time,
    $(D_t\,\boldsymbol{\eta})_l = \boldsymbol{\eta}_{l+1} - \boldsymbol{\eta}_l$.
    Together with the box clipping in Eq.~\eqref{eq:cf-proj}, these
    terms keep the counterfactual close to the original input in both
    magnitude and temporal evolution.

    \item \textbf{Plausibility} is supported by the fourth term:
    $$\lambda_{\text{plaus}}\,\bigl[d_M(X+\boldsymbol{\eta}) - d_M(X)\bigr]_+ .$$
    Here $d_M(\cdot)$ is the Mahalanobis distance from the per-feature
    training mean (using the regularised training covariance). The hinge form penalizes only excess deviation from the training distribution: a test instance that already lies at the periphery of the training cloud is not over-penalized, but any further drift incurs a cost. Plausibility is independently verified after the optimization by an IsolationForest scorer (see
    Section~\ref{sec:eval}).

    \item \textbf{Sparsity}, The fourth CF property is not differentiable and therefore not encoded in $\mathcal{L}_{\text{CF}}$ directly. Instead, we address it \emph{post-hoc} by Growing Spheres feature selection \cite{laugel2018comparison}: editable features are ranked by ascending mean perturbation magnitude $\bar{\Delta}_f = \tfrac{1}{L}\sum_{l=1}^L |\boldsymbol{\eta}_{l,f}|$ and greedily reverted to their original columns. If the counterfactual still meets the target after reverting feature $f$, the reversion is accepted. The final counterfactual modifies only the features whose changes are necessary to keep $\widehat{\text{RUL}}(X') \geq \text{RUL}^\star$.
\end{itemize}

\begin{figure}[t]
\centering
\resizebox{0.82\columnwidth}{!}{%
\begin{tikzpicture}[
  node distance=3mm,
  font=\scriptsize,
  box/.style={draw, rounded corners=1.5pt, align=center, inner sep=2.5pt,
              minimum height=6mm, fill=black!4, line width=0.35pt},
  databox/.style={draw, rounded corners=1.5pt, align=center, inner sep=2.5pt,
                  minimum height=6mm, fill=black!10, line width=0.4pt},
  lstm/.style={draw, rounded corners=1.5pt, align=center, inner sep=3pt,
               minimum height=7mm, minimum width=55mm,
               fill=blue!10, line width=0.4pt},
  prop/.style={draw, rounded corners=1.5pt, align=center, inner sep=2pt,
               minimum height=6mm, minimum width=18mm, line width=0.35pt},
  optbox/.style={draw, rounded corners=1.5pt, align=center, inner sep=2.5pt,
                 minimum height=6mm, fill=olive!22, line width=0.4pt},
  filterbox/.style={draw, rounded corners=1.5pt, align=center, inner sep=2.5pt,
                    minimum height=6mm, fill=red!14, line width=0.4pt},
  outbox/.style={draw, rounded corners=1.5pt, align=center, inner sep=2.5pt,
                 minimum height=6mm, fill=blue!22, line width=0.4pt},
  arrow/.style={-{Latex[length=1.3mm]}, line width=0.35pt},
  trainarrow/.style={-{Latex[length=1.3mm]}, line width=0.35pt, dashed},
  lab/.style={font=\tiny\itshape, gray}
]

\node[lstm] (lstm) at (0,0)
  {Survival LSTM $f_\theta$\\[0.5pt]
   $\hat h_{1:L}=f_\theta(X)$};

\node[box, anchor=north] (window) at ([xshift=-14mm, yshift=-7mm]lstm.south)
  {Window construction\\[0.5pt]
   $\mathcal{W}=\{(X^{(i,k)},t^{(i,k)},\delta^{(i,k)})\}$};
\node[box, anchor=north] (test) at ([xshift=14mm, yshift=-7mm]lstm.south)
  {Test window\\[0.5pt] $X\in\mathbb{R}^{L\times D}$};

\draw[trainarrow] (window.north) -- (window.north |- lstm.south)
  node[midway, lab, anchor=east] {trains\,};
\draw[arrow] (test.north) -- (test.north |- lstm.south)
  node[midway, lab, anchor=west] {\,infer};

\node[databox] (data) at ($(window.south)!0.5!(test.south) + (0,-8mm)$)
  {Dataset $\mathcal{D}=\{(X^{(i)},T^{(i)},\delta^{(i)})\}_{i=1}^{N}$};
\draw[arrow] (data.north) -- ++(0,1.5mm) -| (window.south);
\draw[arrow] (data.north) -- ++(0,1.5mm) -| (test.south);

\node[box, above=4mm of lstm] (rul)
  {Hazards $\hat h_{1:L}$, $\widehat S_{1:L}$, $\widehat{\mathrm{RUL}}(X)$};
\draw[arrow] (lstm.north) -- (rul.south);

\node[prop, fill=olive!22] (val)   at ([xshift=-22mm,yshift=8mm]rul.north)
  {\textbf{Validity}\\ $\mathcal{L}_{\mathrm{rul}}$};
\node[prop, fill=blue!10]  (prox)  at ([yshift=8mm]rul.north)
  {\textbf{Proximity}\\ $\mathcal{L}_{\mathrm{dist}}+\mathcal{L}_{\mathrm{smooth}}$};
\node[prop, fill=red!14]   (plaus) at ([xshift=22mm,yshift=8mm]rul.north)
  {\textbf{Plausibility}\\ $\mathcal{L}_{\mathrm{plaus}}$};

\draw[arrow] (rul.north) -- (val.south);
\draw[arrow] (rul.north) -- (prox.south);
\draw[arrow] (rul.north) -- (plaus.south);

\node[optbox, above=5mm of prox] (adam)
  {Adam optimization\\[0.5pt]
   $Z^{\star}=\arg\min_{Z}\mathcal{L}_{\mathrm{CF}}(Z)$};
\draw[arrow] (val.north)   |- (adam.west);
\draw[arrow] (prox.north)  -- (adam.south);
\draw[arrow] (plaus.north) |- (adam.east);

\node[box, above=4mm of adam] (densecf)
  {Dense counterfactual $X^{\dagger}=X+Z^{\star}$};
\draw[arrow] (adam.north) -- (densecf.south);

\node[filterbox, above=4mm of densecf] (gs)
  {Growing-Spheres feature selection\\[0.5pt]
   keep minimal $\mathcal{F}^{\star}\subseteq\mathcal{F}_{\mathrm{edit}}$};
\draw[arrow] (densecf.north) -- (gs.south);

\node[outbox, above=4mm of gs] (sparse)
  {Sparse counterfactual $X^{\star}$\\[0.5pt]
   $\widehat{\mathrm{RUL}}(X^{\star})\geq\tau$};
\draw[arrow] (gs.north) -- (sparse.south);

\end{tikzpicture}%
}
\caption{SurvCF pipeline.}
\label{fig:survcf}
\end{figure}
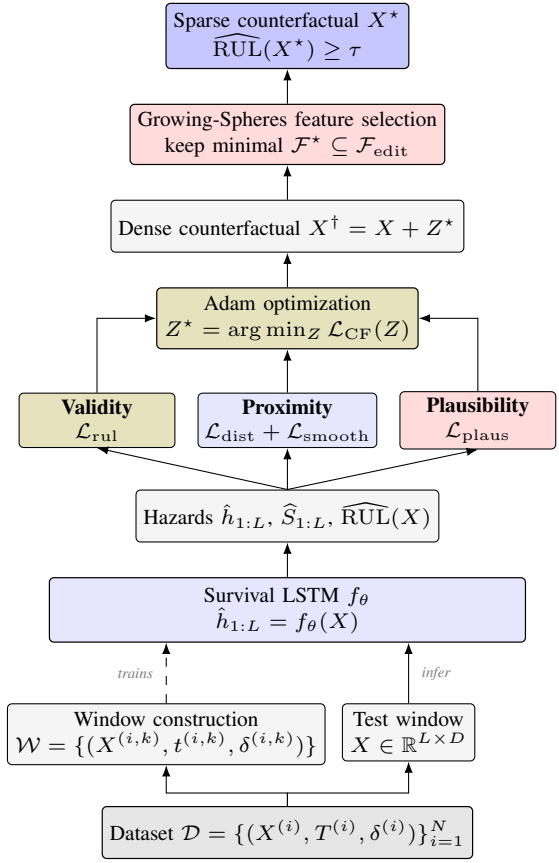

\vspace{2mm}
Fig.~\ref{fig:survcf} gives an end-to-end view of SurvCF. In the lower-left
branch, the dataset $\mathcal{D}$ is sliced into the windowed training set
$\mathcal{W}$ and used to fit the survival LSTM~$f_\theta$ (dashed
``trains'' arrow). Once the model is trained, the lower-right branch passes
a test window $X$ through the \emph{same} $f_\theta$ to produce the hazards
$\hat h_{1:L}$, survival curve $\widehat S_{1:L}$, and predicted RUL. The
upper half of the figure shows the counterfactual stage, in which Adam
optimisation jointly enforces the three CF properties (Validity, Proximity,
Plausibility) to produce a dense counterfactual $X^{\dagger}$, which
Growing-Spheres feature selection then sparsifies into the final $X^{\star}$
satisfying $\widehat{\mathrm{RUL}}(X^{\star})\geq\tau$.

\begin{algorithm}[t]
\caption{Counterfactual Generation}
\label{alg:cf}
\begin{algorithmic}[1]
\Require Model $f_\theta$, window $X$, target $\rho$, editable mask $M_e$,
bounds $(\mathbf{lb}, \mathbf{ub})$, overshoot $\kappa$, steps $S$
\Ensure Counterfactual $X^\star$

\State $\text{RUL}^\star \gets (1+\rho)\,\widehat{\text{RUL}}(X)$;\;
       $\text{RUL}^\star_{\text{int}} \gets \kappa\,\text{RUL}^\star$
\State $Z \gets \arg\min_Z \mathcal{L}_{\text{CF}}(Z)$ via $S$ Adam steps
       \Comment{validity, proximity, plausibility}
\State $X^\star \gets \mathrm{clip}(X + Z \odot M_e,\, \mathbf{lb},\, \mathbf{ub})$

\For{$f \in \mathcal{F}_e$ in ascending order of $\bar{\Delta}_f$}
       \Comment{sparsity}
  \State $X^{\text{trial}} \gets X^\star$ with column $f$ reset to $X_{:,f}$
  \If{$\widehat{\text{RUL}}(X^{\text{trial}}) \geq \text{RUL}^\star$}
    \State $X^\star \gets X^{\text{trial}}$ \Comment{$f$ not needed; revert it to the original}
  \EndIf
\EndFor

\State \Return $X^\star$
\end{algorithmic}
\end{algorithm}

\section{Evaluation Metrics}
\label{sec:eval}

We assess counterfactual quality using four standard properties from the CF literature \cite{pludowski2025mascots,wang2024glacier,guidotti2024counterfactual}, including validity, proximity, sparsity, and plausibility, and complement these with a continuous RUL-improvement measure and per-engine runtime. Let $X$ be the original test window, $X^{\star}$ the counterfactual produced by the pipeline in Section~\ref{sec:cf}, $\widehat{\mathrm{RUL}}(\cdot)$ the trained survival model, and $\rho>0$ the user-specified improvement factor with target $\tau=(1+\rho)\,\widehat{\mathrm{RUL}}(X)$. All metrics are aggregated across the test set as mean~$\pm$~standard deviation.

Next to each metric name, we indicate its desired direction: $\uparrow$ means \emph{higher is better}, so a method performs well when the value is large (e.g.\ Validity, Sparsity, Plausibility, RUL improvement); $\downarrow$ means \emph{lower is better}, so smaller values indicate better performance (e.g.\ Proximity, Runtime).

\vspace{2mm}

\paragraph{Validity ($\uparrow$)} A binary indicator equal to $1$ when the predicted RUL on the counterfactual reaches or exceeds the user-specified target:
\begin{equation}
v(X^{\star}) =
\begin{cases}
1 & \text{if } \widehat{\mathrm{RUL}}(X^{\star}) \geq \tau, \\
0 & \text{otherwise.}
\end{cases}
\end{equation}
The \emph{validity rate} of a method is the mean of $v(X^{\star})$
across the test set.

\paragraph{Proximity ($\downarrow$)}
The Euclidean distance between original and counterfactual windows in scaled feature space, normalized by the number of cells:
\begin{equation}
\mathrm{Prox}(X^{\star}, X)
= \tfrac{1}{L\,D}\,\lVert X^{\star} - X \rVert_{2},
\end{equation}
where $\lVert\cdot\rVert_{2}$ denotes the $L_2$ norm of the perturbation flattened to an $LD$-dimensional vector (equivalently, the Frobenius norm of the $L\times D$ matrix $X^{\star}-X$).

\paragraph{Sparsity ($\uparrow$).}
Sparsity is computed over the editable feature set $\mathcal{F}_e$. Let $\mathcal{F}^{\star}\subseteq\mathcal{F}_e$
denote the subset of editable features that the final counterfactual
$X^{\star}$ actually modifies, i.e.\ the features retained after the
Growing-Spheres pruning step. We report the fraction of editable
features the counterfactual leaves untouched:
\begin{equation}
\mathrm{Sp}(X^{\star}, X) \;=\; 1 - \frac{|\mathcal{F}^{\star}|}{|\mathcal{F}_e|}.
\end{equation}
The metric lies in $[0,1]$; higher values indicate that the
counterfactual concentrates its action on fewer actuators and is
therefore more interpretable for a maintenance operator.


\paragraph{Plausibility ($\uparrow$)}
A binary indicator from the IsolationForest decision function $\Psi$,
fitted on the per-feature temporal means of the training windows
restricted to editable features:
\begin{equation}
\mathrm{Pl}(X^{\star}) =
\begin{cases}
1 & \text{if } \Psi\!\bigl(\bar{X}^{\star}_{\mathcal{F}_e}\bigr) > 0, \\
0 & \text{otherwise,}
\end{cases}
\end{equation}
where $\bar{X}^{\star}_{\mathcal{F}_e}$ is the per-feature temporal mean
of $X^{\star}$, restricted to editable features. We additionally report
the percentile of $\Psi\!\bigl(\bar{X}^{\star}_{\mathcal{F}_e}\bigr)$
within the training-window score distribution, which quantifies
\emph{how} plausible the CF is.


\paragraph{RUL improvement (\%) ($\uparrow$)}
The relative change in predicted RUL between original and counterfactual:
\begin{equation}
\Delta\mathrm{RUL}_{\%}
= 100 \cdot
  \frac{\widehat{\mathrm{RUL}}(X^{\star}) - \widehat{\mathrm{RUL}}(X)}
       {\widehat{\mathrm{RUL}}(X)}.
\end{equation}

\paragraph{Runtime ($\downarrow$)}
Wall-clock seconds per engine to generate a counterfactual, averaged
across the test set.

\section{Experiments}
\label{sec:experiments}

We evaluate the proposed SurvCF($t$) framework on three publicly available run-to-failure benchmarks in PdM. Each benchmark exposes a multivariate sensor stream observed up to either a recorded failure or an end-of-study censoring time. The experimental question is the same across all three: given the last observation window before a maintenance decision is required, can a counterfactual perturbation of the actionable channels extend the predicted RUL of the asset, and at what cost to the four CF properties of validity, proximity, sparsity and plausibility introduced in Section~\ref{sec:eval}? The next subsections present per-dataset results across a sweep of RUL-improvement targets, followed by an editable-set ablation that quantifies the trade-off between operational realism and CF capacity.

\subsection{Datasets}
\label{sec:datasets}


\subsubsection{C-MAPSS}
\label{sec:dataset_cmapss}
We evaluate the proposed method on the Commercial Modular Aero-Propulsion
System Simulation (C-MAPSS) turbofan engine degradation benchmark released by NASA~\cite{saxena2008damage}. C-MAPSS provides four sub-datasets (\textsc{FD001} to \textsc{FD004}), each containing the multivariate sensor histories of a fleet of simulated turbofan engines run from a healthy state to failure.
Each engine i produces a time series of 
24 channels, three operating settings, and 21 simulated sensor measurements. In this experiment, the \textsc{FD001} subset is used, and the readouts are artificially censored at cycle=250 to make them compatible with a survival analysis framework. For model inputs, we use windows of length w=20 with time-step dt=5 (i.e., the model ingests twenty consecutive cycles and predicts conditional hazards at 5‑cycle resolution).


\subsubsection{N-CMAPSS}

The New Commercial Modular Aero-Propulsion System Simulation (N-CMAPSS)
benchmark, released by NASA and ETH~Z\"urich~\cite{arias2021aircraft},
extends the original C-MAPSS dataset with run-to-failure trajectories of
turbofan engines simulated under \emph{real flight envelopes} recorded
from commercial operations rather than synthetic operating points. Each
engine is simulated at a $1$\,Hz sampling rate over its entire useful
life, producing trajectories that span tens of complete flights
including climb, cruise, and descent segments. The dataset is
distributed in eight HDF5 sub-files (\textsc{DS01}--\textsc{DS08}),
each grouping of engines that share a common flight class and degradation
profile. Every HDF5 file partitions its engines into a \emph{dev} split
(training/validation) and a held-out \emph{test} split.
Each engine record contains four groups of variables: scenario
descriptors $W$ (flight envelope inputs), measured physical sensors
$X_s$, auxiliary metadata $A$ (unit ID, cycle index, flight class), and the per-cycle RUL $Y$.
In our SurvCF(t) pipeline, we first aggregate the original per‑second recordings to per‑cycle summaries (per‑cycle means) and merge the selected subsets (we used DS01–DS03 for reported experiments). Samples for the survival model are produced as fixed‑length windows (we found window L=20, with dt=2, to be the best tradeoff), and labels / censoring information are prepared in the standard survival format with censoring at cycle=65.




\subsubsection{\textbf{Scania Component\_X (the main case study)}}
\label{sec:dataset_compx}

The third benchmark is a real-world dataset called the Component\_X dataset~\cite{kharazian2025scania}, released by Scania AB. This data contains a fleet-level run-to-failure record of more than $33{,}000$ heavy-duty trucks. Each vehicle contributes a sequence of readouts, where each readout reports two kinds of variables: (i)~scalar counters, which monotonically accumulate vehicle activity, and (ii) histogram variables, which distribute the same physical quantity across operating-condition bins. Survival labels are taken from the companion \texttt{train\_tte.csv} file: the binary indicator \texttt{in\_study\_repair} encodes whether a Component\_X failure event was observed during the study, and \texttt{length\_of\_study\_time\_step} gives the corresponding failure or right-censoring time.
From the raw readouts we retain (a) non-histogram
counter columns and (b)~the $97$ histogram bin columns belonging to the six histogram groups. Because the eight counter columns are cumulative rather than per-interval, we convert each one into a time series of \emph{per readout deltas} by subtracting the previous value within the same vehicle,
\begin{equation}
x^{(d)}_{v,t} \;=\; x_{v,t} - x_{v,t-1},
\qquad
x^{(d)}_{v,t_0}=0,
\end{equation}
so that the differenced channel reports the activity \emph{added since the last readout}. 
All input channels are standardized using the training‑fold mean and standard deviation. Sliding windows of length $L=20$ are extracted, and the remaining time‑to‑failure is discretized with $\Delta t=10$ for the discrete‑time survival target.

It is worth mentioning that the previous two datasets were generated with fixed sampling intervals, so each readout corresponds to one \texttt{time\_step}. Component~X (like most industrial datasets), however, has non-uniform (irregular) timestamps: intervals between consecutive readouts vary across units. SurvCF($t$) accommodates this natively: the survival LSTM, the gradient-CF optimizer, and the Growing-Spheres and plausibility stages all operate on \emph{readout sequences} of fixed length $L$ rather than on uniform time grids. The window of $L=20$ readouts used at decision time is therefore a sequence-length constraint, not a wall-clock constraint (i.e., $L=20$ refers to the last $20$ readouts of a vehicle, not to a span of $20$ \texttt{time\_step} units). For an irregularly sampled vehicle, the same $20$ readouts may cover, for instance, about $100$ \texttt{time\_step} units. To allow the model to condition on the real-time spacing between readouts, we additionally augment the input with a per-vehicle \texttt{gap} feature $g_{v,t}=t-t_{v,t-1}$, exposing the inter-readout interval as a covariate. This design lets SurvCF($t$) treat regular and irregular sampling regimes within a single framework, rather than requiring resampling or interpolation as a preprocessing step.

\section{Results}

\subsection{Result for C-MAPSS}

Table~\ref{tab:cmapss_sweep} reports SurvCF's behavior on C‑MAPSS across five RUL‑improvement targets. The method attains its highest validity at the smallest target ($+5\%$: $0.733$), and shows moderate validity at $+10\%$ and $+20\%$ ($0.533$ each), rising to $0.6$ at $+30\%$ and $+50\%$. Counterfactuals remain very proximate to the originals, with proximity scores of $0.008$, $0.009$, and $0.013$ at $+5\%$, $+10\%$, and $+20\%$ (and $0.015$, $0.0187$ at $+30\%$ and $+50\%$). Sparsity is favorable $0.722$, $0.711$, and $0.544$ for the $+5\%$–$+20\%$ targets (falling to $0.444$ and $0.3$ at larger targets). This indicates compact, interpretable edits. Plausibility remains within realistic operating regions (e.g., $0.267$ at $+5\%$ up to $0.533$ at the larger targets). Runtime is stable and fast, ranging roughly $1.24$–$1.32$ s (for example, $1.316$ s at $+5\%$ and $1.280$ s at $+50\%$), so SurvCF is both practical and efficient for C‑MAPSS evaluations.

\begin{table}[!htbp]
\centering
\scriptsize
\setlength{\tabcolsep}{2.5pt}
\renewcommand{\arraystretch}{1.08}
\caption{SurvCF on C-MAPSS, across five RUL-improvement targets. Values are mean $\pm$ std. Arrows indicate direction of improvement.}
\label{tab:cmapss_sweep}
\begin{tabular}{@{}cccccc@{}}
\toprule
\textbf{Target} & \textbf{Val.} $\uparrow$ & \textbf{Prox.} $\downarrow$ & \textbf{Spars.} $\uparrow$ & \textbf{Plaus.} $\uparrow$ & \textbf{Runtime (s)} $\downarrow$ \\
\midrule
+5\%  & $0.733_{\scriptscriptstyle \pm 0.458}$ & $0.008_{\scriptscriptstyle \pm 0.003}$ & $0.722_{\scriptscriptstyle \pm 0.224}$ & $0.267_{\scriptscriptstyle \pm 0.458}$ & $1.316_{\scriptscriptstyle \pm 0.221}$ \\
+10\% & $0.533_{\scriptscriptstyle \pm 0.516}$ & $0.009_{\scriptscriptstyle \pm 0.003}$ & $0.711_{\scriptscriptstyle \pm 0.213}$ & $0.200_{\scriptscriptstyle \pm 0.414}$ & $1.243_{\scriptscriptstyle \pm 0.007}$ \\
+20\% & $0.533_{\scriptscriptstyle \pm 0.516}$ & $0.013_{\scriptscriptstyle \pm 0.004}$ & $0.544_{\scriptscriptstyle \pm 0.183}$ & $0.200_{\scriptscriptstyle \pm 0.414}$ & $1.251_{\scriptscriptstyle \pm 0.007}$ \\
+30\% & $0.600_{\scriptscriptstyle \pm 0.507}$ & $0.015_{\scriptscriptstyle \pm 0.005}$ & $0.444_{\scriptscriptstyle \pm 0.185}$ & $0.533_{\scriptscriptstyle \pm 0.516}$ & $1.265_{\scriptscriptstyle \pm 0.038}$ \\
+50\% & $0.600_{\scriptscriptstyle \pm 0.507}$ & $0.018_{\scriptscriptstyle \pm 0.007}$ & $0.300_{\scriptscriptstyle \pm 0.191}$ & $0.533_{\scriptscriptstyle \pm 0.516}$ & $1.280_{\scriptscriptstyle \pm 0.027}$ \\
\bottomrule
\end{tabular}
\end{table}

\subsection{Result for the N-CMAPSS} \label{sec:results_ncmapss}

Table~\ref{tab:ncmapss_sweep} reports SurvCF's behavior on N-CMAPSS across five RUL-improvement targets. The method shows strong practical utility: validity is $0.923$ at both $+5\%$ and $+10\%$, and remains $0.769$ at $+20\%$, indicating that SurvCF can reliably generate valid counterfactuals for moderate target increases. At the same time, the counterfactuals remain close to the original trajectories, with proximity scores of $0.011$, $0.016$, and $0.030$ at $+5\%$, $+10\%$, and $+20\%$, respectively. The sparsity values are similarly favorable, with $0.275$, $0.181$, and $0.159$ at these same targets, showing that SurvCF achieves its edits with relatively few feature changes.
Plausibility is also consistently high, reaching $0.769$ at $+5\%$ and $+10\%$, $0.923$ at $+20\%$ and $+30\%$, and $0.846$ at $+50\%$. 
It shows that the Mahalanobis penalty in the CF loss and the Growing-Spheres sparsification step jointly ensure that the counterfactuals remain within the training distribution.
Runtime, also remains stable between $1.848$ s and $1.997$ s, indicating that SurvCF remains efficient across all evaluated targets. The poorer results at higher targets stem from the fact that those improvements require unrealistically large or out‑of‑distribution edits, which SurvCF prevents to preserve sparsity and plausibility.

\begin{table}[!htbp]
\centering
\scriptsize
\setlength{\tabcolsep}{2.5pt}
\renewcommand{\arraystretch}{1.05}
\caption{SurvCF on N-CMAPSS (DS01+DS02+DS03). Values are mean $\pm$ std. Arrows indicate direction of improvement.}
\label{tab:ncmapss_sweep}
\begin{tabular}{@{}cccccc@{}}
\toprule
\textbf{Target} & \textbf{Val.} $\uparrow$ & \textbf{Prox.} $\downarrow$ & \textbf{Spars.} $\uparrow$ & \textbf{Plaus.} $\uparrow$ & \textbf{Runtime (s)} $\downarrow$ \\
\midrule
+5\%  & $0.923_{\scriptscriptstyle \pm 0.28}$ & $0.011_{\scriptscriptstyle \pm 0.001}$ & $0.275_{\scriptscriptstyle \pm 0.13}$ & $0.769_{\scriptscriptstyle \pm 0.44}$ & $1.997_{\scriptscriptstyle \pm 0.218}$ \\
+10\% & $0.923_{\scriptscriptstyle \pm 0.28}$ & $0.016_{\scriptscriptstyle \pm 0.002}$ & $0.181_{\scriptscriptstyle \pm 0.09}$ & $0.769_{\scriptscriptstyle \pm 0.44}$ & $1.941_{\scriptscriptstyle \pm 0.025}$ \\
+20\% & $0.769_{\scriptscriptstyle \pm 0.44}$ & $0.030_{\scriptscriptstyle \pm 0.004}$ & $0.159_{\scriptscriptstyle \pm 0.07}$ & $0.923_{\scriptscriptstyle \pm 0.28}$ & $1.969_{\scriptscriptstyle \pm 0.081}$ \\
+30\% & $0.077_{\scriptscriptstyle \pm 0.28}$ & $0.041_{\scriptscriptstyle \pm 0.004}$ & $0.121_{\scriptscriptstyle \pm 0.10}$ & $0.923_{\scriptscriptstyle \pm 0.28}$ & $1.931_{\scriptscriptstyle \pm 0.017}$ \\
+50\% & $0.000_{\scriptscriptstyle \pm 0.00}$ & $0.046_{\scriptscriptstyle \pm 0.007}$ & $0.000_{\scriptscriptstyle \pm 0.00}$ & $0.846_{\scriptscriptstyle \pm 0.38}$ & $1.848_{\scriptscriptstyle \pm 0.006}$ \\
\bottomrule
\end{tabular}
\end{table}

\subsection{Results for Component X (Our Case Study)}

Table~\ref{tab:cmapss_sweep_onecol} reports SurvCF's behavior on the Component\_X benchmark across five RUL‑improvement targets. SurvCF generated valid counterfactuals for 37.9\% of cases at $+5\%$, 27.3\% at $+10\%$, 18.2\% at $+20\%$, 16.7\% at $+30\%$, and 15.2\% at $+50\%$. The counterfactuals remain extremely proximate to the originals (proximity $\approx 0.003$ for all targets). 

Feature-level $L_0$ sparsity declines from $15.9\%$ at $+5\%$ to $5.7\%$ at $+50\%$, meaning that the average counterfactual touches between $84\%$ and $94\%$ of editable counters. Growing-Spheres pruning has limited room to operate. This is consistent with the operational structure of the Component~X
data: the editable counters track distinct but interdependent
vehicle-activity modes, so no single counter dominates the survival model's RUL response, and the optimizer is forced to recruit nearly the entire editable set even for small perturbations.



Plausibility stays high across targets (score $\approx 0.955$), showing the edits lie within realistic operating regions. Runtime is stable and low (approximately 1.80 s, 1.79 s, 1.75 s, 1.74 s, and 1.76 s for $+5\%$–$+50\%$), demonstrating SurvCF is practical and efficient on Component\_X.

\begin{table}[!htbp]
\centering
\scriptsize
\setlength{\tabcolsep}{2.0pt}
\renewcommand{\arraystretch}{1.02}
\caption{SurvCF on Component X data. Values are mean $\pm$ std.}
\label{tab:cmapss_sweep_onecol}
\begin{tabular}{@{}lccccc@{}}
\toprule
\textbf{Target} & \textbf{Val.} $\uparrow$ & \textbf{Prox.} $\downarrow$ & \textbf{Spars. L$_0$} $\uparrow$ & \textbf{Plaus.} $\uparrow$ & \textbf{Runtime (s)} $\downarrow$ \\
\midrule
+5\%  & $0.379_{\scriptscriptstyle\pm\,0.489}$ & $0.003_{\scriptscriptstyle\pm\,0.002}$ & $0.159_{\scriptscriptstyle\pm\,0.255}$ & $0.955_{\scriptscriptstyle\pm\,0.210}$ & $1.801_{\scriptscriptstyle\pm\,0.100}$ \\
+10\% & $0.273_{\scriptscriptstyle\pm\,0.449}$ & $0.003_{\scriptscriptstyle\pm\,0.002}$ & $0.117_{\scriptscriptstyle\pm\,0.223}$ & $0.955_{\scriptscriptstyle\pm\,0.210}$ & $1.794_{\scriptscriptstyle\pm\,0.046}$ \\
+20\% & $0.182_{\scriptscriptstyle\pm\,0.389}$ & $0.003_{\scriptscriptstyle\pm\,0.002}$ & $0.078_{\scriptscriptstyle\pm\,0.182}$ & $0.955_{\scriptscriptstyle\pm\,0.210}$ & $1.754_{\scriptscriptstyle\pm\,0.058}$ \\
+30\% & $0.167_{\scriptscriptstyle\pm\,0.376}$ & $0.003_{\scriptscriptstyle\pm\,0.002}$ & $0.072_{\scriptscriptstyle\pm\,0.178}$ & $0.955_{\scriptscriptstyle\pm\,0.210}$ & $1.739_{\scriptscriptstyle\pm\,0.034}$ \\
+50\% & $0.152_{\scriptscriptstyle\pm\,0.361}$ & $0.003_{\scriptscriptstyle\pm\,0.002}$ & $0.057_{\scriptscriptstyle\pm\,0.151}$ & $0.955_{\scriptscriptstyle\pm\,0.210}$ & $1.764_{\scriptscriptstyle\pm\,0.059}$ \\
\bottomrule
\end{tabular}
\end{table}

\subsubsection{\textbf{Qualitative example (Vehicle~18655)}} 
\begin{figure}[htbp]
\centering
\includegraphics[width=1\columnwidth]{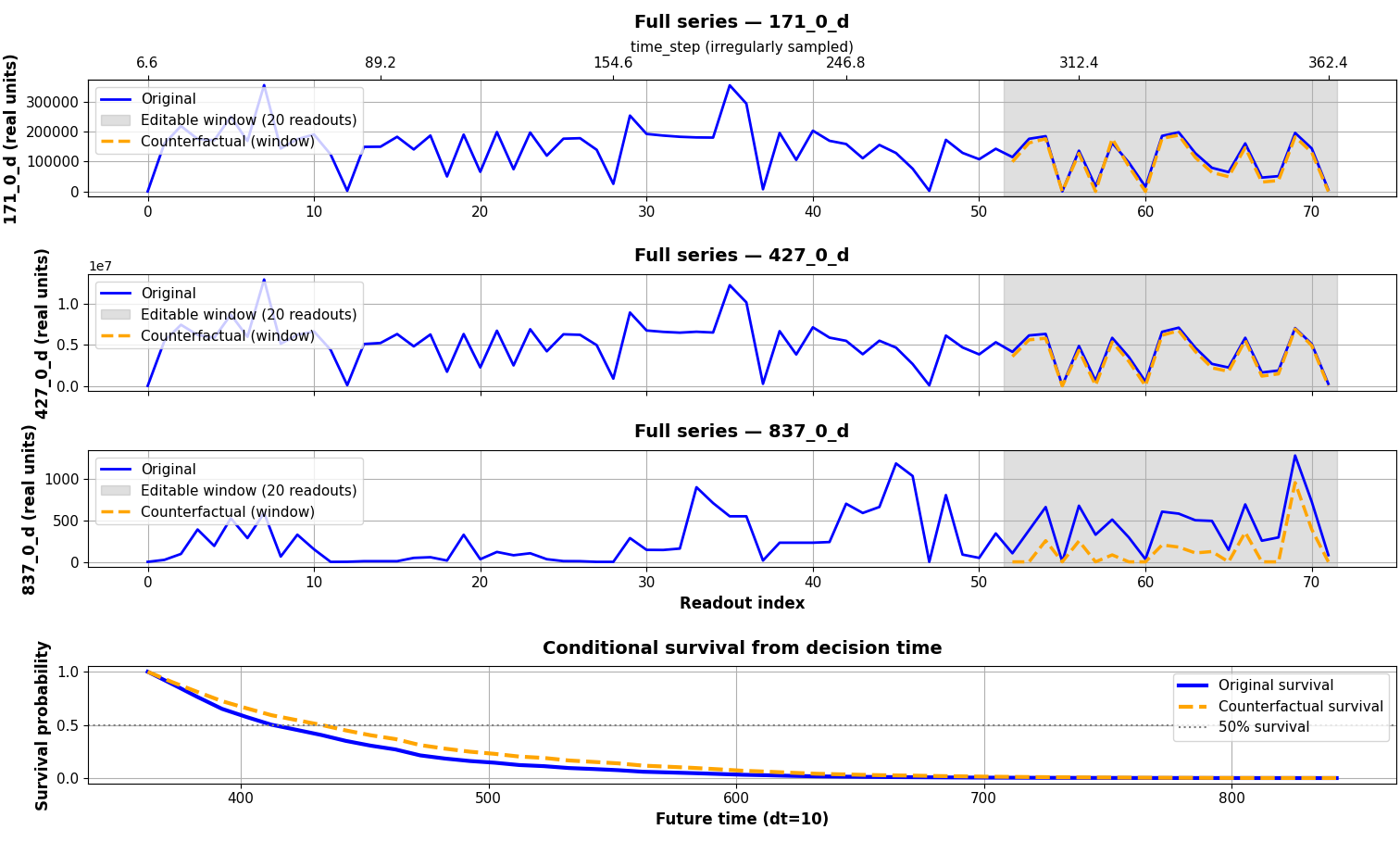}
\caption{case study sample. test vehicle \#18655. RUL\_before = 77.14, RUL\_after = 92.52}
\label{case_study}
\end{figure}

Fig.~\ref{case_study} shows the SurvCF($t$) pipeline applied to test vehicle~18655 at a $+20\%$ RUL-improvement target.
The trained survival LSTM assigns the original last-$L{=}20$-readout window a predicted RUL of $77.14$\,cycles; the target is therefore $1.20 \times 77.14 = 92.57$\,cycles. After Step~1 (gradient CF), Step~2 (Growing-Spheres feature pruning), Step~3 (plausibility scoring), SurvCF returns a counterfactual that \textbf{(i)}~reaches a predicted RUL of $96.97$\,cycles, exceeding the target by $4.40$\,cycles and yielding a relative improvement of $+25.7\%$ (Validity $=1$); \textbf{(ii)}~modifies only $3$ of the $8$ editable activity counters ($171\_0\_d$, $427\_0\_d$, and $837\_0\_d$), corresponding to a feature-level $L_0$ sparsity of $0.625$; \textbf{(iii)}~achieves this with an average per-cell $L_2$ perturbation of $0.0004$ in the scaled feature space (Proximity); and \textbf{(iv)}~remains in the training distribution as flagged by the IsolationForest plausibility check, with a decision-function score of $+0.184$, placing the counterfactual at the $83.9$\textsuperscript{th} percentile of the training-window score distribution (Plausibility $=1$). The top three panels of Fig.~\ref{case_study} show the original (blue) and counterfactual (orange dashed) trajectories of the three modified counters across the truck's full readout history, with the editable window of $L{=}20$ readouts marked in grey. The bottom panel compares the conditional survival curves at the decision time, in which the counterfactual curve sits above the original and crosses $50\%$ survival approximately $30$\, cycles later. Vehicle~18655 thus illustrates that SurvCF($t$) is capable, on real censored industrial data, of producing a counterfactual that is simultaneously \emph{valid}, \emph{sparse}, \emph{small in magnitude}, and \emph{plausible} relative to the training distribution that are the four desiderata commonly used to evaluate counterfactual explanations.

\begin{table}[ht]
\centering
\caption{Counterfactual properties for truck \#18655.}
\label{tab:cf_properties_example}
\begin{tabular}{@{} l l @{}}
\toprule
Property & Value \\
\midrule
Target & +20\% RUL improvement (77.14 $\rightarrow$ 92.57) \\
Validity & 1 (predicted RUL = 96.97, PASS) \\
Proximity  & 0.0004  \\
Sparsity (L0, editable) & 0.625 (3/8 features modified) \\
Plausibility & 1  \\
RUL improvement & +25.7\% (77.14 $\rightarrow$ 96.97) \\
\bottomrule
\end{tabular}
\end{table}



\section{Conclusion}\label{sec:Conclusion}






We have introduced \textbf{SurvCF($t$)}\footnote{Source code:
\url{https://github.com/XaraKar/SurvCF}}, a counterfactual-generation framework for survival models trained on multivariate, censored time series. The method (i) trains a discrete-time hazard LSTM on windowed run-to-failure data, (ii) generates dense gradient-based counterfactuals against a composite loss balancing the four CF properties, (iii) sparsifies the result via a Growing-Spheres feature-selection step, and (iv) reports each counterfactual against an IsolationForest plausibility model fitted on the training-window distribution.

\paragraph{Contributions}

To our knowledge, \textit{SurvCF(t)} is the first counterfactual framework that combines (a) survival modeling under right‑censoring, (b) multivariate time‑series inputs typical of industrial sensor streams, and (c) supports joint validity/proximity/sparsity/plausibility counterfactual explanation. Existing tabular CF methods cannot accommodate the temporal structure or the censored survival objective. Also, existing time‑series CF methods (which address classification or forecasting) do not optimize for survival outcomes, and prior work that mixes counterfactuals with survival analysis has focused on static/tabular covariates only, rather than multivariate, possibly irregularly sampled sensor windows. SurvCF therefore fills a previously unaddressed niche by producing actionable, interpretable counterfactuals for time‑to‑event prediction on real industrial telemetry, enabling operator‑centric recommendations that respect both temporal dynamics and censoring. 

\paragraph{Practical relevance}
The framework targets a real industrial need: censored, multivariate sensor data is abundant in predictive-maintenance settings, and
operators currently lack tools that turn RUL predictions into
operationally meaningful actions. SurvCF directly provides such
recommendations, identifying for each asset the minimal, plausible,
and (where the editable set is restricted appropriately) operator-
actionable perturbation that would extend its predicted remaining
useful life.

\paragraph{Empirical findings}
Across three benchmarks SurvCF reaches a validity of $0.53$
($+45.3\%$ RUL gain at the $+50\%$ target on C-MAPSS), $0.92$
($+5\%$ and $+10\%$ targets on N-CMAPSS), and up to $0.38$ on
Component~X, with plausibility consistently above $0.77$ across all
configurations. An editable-set ablation on N-CMAPSS quantifies the
trade-off between operational realism and CF capacity. The
counterfactuals are sparse (typically modifying a small fraction of
editable features), small in per-cell magnitude, and produced in
$1$--$2$ seconds per asset on a CPU.

\paragraph{Future work}
Future work can investigate the 
extension of SurvCF to multi-modal datasets that include more modalities than time series data.
Moreover, integration of a user-in-the-loop refinement step that allows domain experts to select among candidate counterfactuals, closing the loop between algorithmic recommendation and operational decision.

\section*{Acknowledgment}

This work has been funded by the Vinnova Program for Transport and mobility solutions (FFI) and partly by Traton AB.

\bibliographystyle{ieeetr}
\bibliography{references}

\end{document}